\documentclass[11pt,a4paper]{article}

\usepackage[hyperref]{naaclhlt2019}

\usepackage{times}
\usepackage{latexsym}
\usepackage{todonotes}
\usepackage{url}
\usepackage{multirow}
\usepackage{todonotes}
\usepackage{booktabs}

\aclfinalcopy 

\title{Predicting the Type and Target of Offensive Posts in Social Media}

\author{Marcos Zampieri\textsuperscript{1}, Shervin Malmasi\textsuperscript{2}, Preslav Nakov\textsuperscript{3}, Sara Rosenthal\textsuperscript{4} \\ \bf{Noura Farra\textsuperscript{5}, Ritesh Kumar\textsuperscript{6}} \\
  \textsuperscript{1}University of Wolverhampton, UK, \textsuperscript{2}Harvard Medical School, USA\\ \textsuperscript{3}Qatar Computing Research Institute, HBKU, Qatar, \textsuperscript{4}IBM Research, USA \\ \textsuperscript{5}Columbia University, USA, \textsuperscript{6}Bhim Rao Ambedkar University, India \\
  {\tt m.zampieri@wlv.ac.uk} \\}

\date{}

\begin{document}
\maketitle

\begin{abstract}
 As offensive content has become pervasive in social media, there has been much research in identifying potentially offensive messages. However, previous work on this topic did not consider the problem as a whole, but rather focused on detecting very specific types of offensive content, e.g., hate speech, cyberbulling, or cyber-aggression. In contrast, here we target several different kinds of offensive content. In particular, we model the task hierarchically, identifying the type and the target of offensive messages in social media. For this purpose, we complied the Offensive Language Identification Dataset (OLID), a new dataset with tweets annotated for offensive content using a fine-grained three-layer annotation scheme, which we make publicly available. We discuss the main similarities and differences between OLID and pre-existing datasets for hate speech identification, aggression detection, and similar tasks. We further experiment with and we compare the performance of different machine learning models on OLID.
\end{abstract}

\section{Introduction}

Offensive content has become pervasive in social media and thus a serious concern for government organizations, online communities, and social media platforms. One of the most common strategies to tackle the problem is to train systems capable of recognizing offensive content, which can then be deleted or set aside for human moderation. In the last few years, there have been several studies on the application of computational methods to deal with this problem.
Prior work has studied offensive language in Twitter \cite{xu2012learning,burnap2015cyber, davidson2017automated, wiegand2018overview}, Wikipedia comments,\footnote{\url{http://bit.ly/2FhLMVz}} and Facebook posts \cite{kumar2018benchmarking}.

\noindent Previous studies have looked into different aspects of offensive language such as the use of abusive language \cite{nobata2016abusive,mubarak2017}, (cyber-)aggression \cite{kumar2018benchmarking}, (cyber-)bullying \cite{xu2012learning,dadvar2013improving}, toxic comments$^1$, hate speech \cite{kwok2013locate,djuric2015hate,burnap2015cyber,davidson2017automated,malmasi2017detecting,malmasi2018}, and offensive language \cite{wiegand2018overview}. 

Recently, \newcite{waseem2017understanding} analyzed the similarities between different approaches proposed in previous work and argued that there was a need for a typology that differentiates between whether the (abusive) language is directed towards a specific individual or entity, or towards a generalized group, and whether the abusive content is explicit or implicit. \newcite{wiegand2018overview} further applied this idea to German tweets. They experimented with a task on detecting offensive vs. non-offensive tweets, and also with a second task on further sub-classifying the offensive tweets as profanity, insult, or abuse. However, to the best of our knowledge, no prior work has explored the target of the offensive language, which might be important in many scenarios, e.g.,~when studying hate speech with respect to a specific target. Below, we aim at bridging this gap.

More generally, in this paper, we expand on the above ideas by proposing a novel three-level hierarchical annotation schema that encompasses the following three general categories:

\begin{itemize}
    \item[\bf A:] Offensive Language Detection
    \item[\bf B:] Categorization of Offensive Language
    \item[\bf C:] Offensive Language Target Identification
\end{itemize}

\begin{table*}[tbh]
\centering
\scalebox{0.92}{
\begin{tabular}{p{12.5cm}ccc}
\toprule
 \bf Tweet & \bf A & \bf B & \bf C \\
\midrule
 @USER He is so generous with his offers. & NOT & --- &  --- \\
 IM FREEEEE!!!! WORST EXPERIENCE OF MY FUCKING LIFE & OFF & UNT & --- \\
 @USER Fuk this fat cock sucker & OFF & TIN & IND \\
 @USER Figures! What is wrong with these idiots? Thank God for @USER & OFF & TIN & GRP \\
\bottomrule
\end{tabular}
}
\caption{Four tweets from the dataset, with their labels for each level of the annotation schema.}
\label{T:examples}
\end{table*}

\newpage
\noindent We further use the above schema to annotate a large dataset of English tweets, which we make publicly available online.\footnote{The data can be downloaded from the following address:\\ \url{http://scholar.harvard.edu/malmasi/olid}} 

The key contributions of this paper can be summarized as follows:

\begin{itemize}
\item We propose a new three-level hierarchical annotation schema for abusive language detection and characterization.
\item We apply the schema to create \textit{Offensive Language Identification Dataset (OLID)}, a new large-scale dataset of English tweets with high-quality annotation of the target and type of offenses.
\item We perform experiments on OLID using different machine learning models for each level of the annotation, thus setting important baselines to compare to in future work.
\end{itemize}

\noindent While each of these sub-tasks tackles a particular type of abuse or offense, they share similar properties and the hierarchical annotation model proposed in this paper aims to capture this. Considering that, for example, an insult targeted at an individual is commonly known as cyberbulling and that insults targeted at a group are known as hate speech, we believe that OLID's use of a hierarchical annotation schema makes it a useful resource for various offensive language identification and characterization tasks.

\section{Hierarchically Modelling Offensive Content}

In the OLID dataset, we use a hierarchical annotation schema split into three levels to distinguish between whether the language is offensive or not (A), its type (B), and its target (C).
Each level is described in more detail in the following subsections and examples are shown in Table~\ref{T:examples}.

\subsection{Level A: Offensive language Detection}

Level A discriminates between the following types of tweets:

\begin{itemize}

\item {\bf Not Offensive (NOT):}
 Posts that do not contain offense or profanity; 

\item {\bf  Offensive (OFF):} Posts containing any form of non-acceptable language (profanity) or a targeted offense, which can be veiled or direct. This includes insults, threats, and posts containing profane language or swear words.

\end{itemize}

\subsection{Level B: Categorization of Offensive Language}

Level B categorizes the type of offense:

\begin{itemize}

\item {\bf Targeted Insult (TIN):} Posts containing insult/threat to an individual, a group, or others;

\item {\bf Untargeted (UNT):} Posts containing non-targeted profanity and swearing. Posts with general profanity are not targeted, but they contain non-acceptable language.

\end{itemize}

\subsection{Level C: Offensive Language Target Identification}

Level C categorizes the targets of insults/threats:

\begin{itemize}

\item {\bf Individual (IND):} Posts targeting an individual. This can be a famous person, a named individual or an unnamed participant in the conversation. Insults and threats targeted at individuals are often defined as cyberbulling. 

\item {\bf Group (GRP):} Posts targeting a group of people considered as a unity due to the same ethnicity, gender or sexual orientation, political affiliation, religious belief, or other common characteristic. Many of the insults and threats targeted at a group correspond to what is commonly understood as hate speech.

\item {\bf Other (OTH)} The target of these offensive posts does not belong to any of the previous two categories (e.g., an organization, a situation, an event, or an issue). 

\end{itemize}

\begin{table}[!ht]
\centering
\scalebox{0.95}{
\begin{tabular}{lc}
\toprule
\bf Keyword & \bf Offensive \% \\
\midrule
medical marijuana & 0.0 \\
they are & 5.9 \\
to:NewYorker & 8.3 \\ \hline
you are & 21.0 \\
she is & 26.6 \\
to:BreitBartNews & 31.6 \\
he is & 32.4 \\
gun control & 34.7 \\
-filter:safe & 58.9 \\ 
conservatives & 23.2 \\
antifa & 26.7 \\
MAGA & 27.7 \\
liberals & 38.0 \\
\bottomrule
\end{tabular}
}
\caption{The keywords from the full dataset (except for the first three rows) and the percentage of offensive tweets for each keyword.}
\label{T:keywords}
\end{table}

\section{Data Collection}

We retrieved the examples in OLID from Twitter using its API and searching for keywords and constructions that are often included in offensive messages, such as `she is' or `to:BreitBartNews'\footnote{\emph{to} is a special Twitter API word indicating that the tweet was directed at a specific account (e.g.,~BreitBartNews).}. The full list of keywords we used is shown in Table~\ref{T:keywords}.

We first carried out a round of trial annotation of 300 instances with six experts using nine keywords. The goal of the trial annotation was (\emph{i})~to evaluate the proposed tagset, (\emph{ii})~to evaluate the data retrieval method, and (\emph{iii})~to create a gold standard with instances that could be used as test questions 
to ensure the quality of the annotators for the rest of the data, which was carried out using crowdsourcing. The keywords used in the trial annotation are shown in the first nine rows of Table~\ref{T:keywords}. We included left (\textit{@NewYorker}) and far-right (\textit{@BreitBartNews}) news accounts because there tends to be political offense in the comments for such accounts. The keyword that resulted in the highest concentration of offensive content was the Twitter `safe' filter, corresponding to tweets that were flagged as unsafe by Twitter (the `-' symbiol indicates `not safe').

Since the vast majority of content on Twitter is not offensive, we tried different strategies to keep the distribution of offensive tweets at around 30\% of the dataset. We excluded some keywords that were not high in offensive content during the trial annotation such as `they are' and `to:NewYorker'. 

\noindent Although `he is' was poor in offensive content in the trial dataset (15\%), we kept it as a keyword in order to avoid gender bias, and we found that in the full dataset it was more offensive (32.4\%). The trial keywords that we ultimately decided to exclude due to low percentage of offensive tweets are shown in the top portion of Table~\ref{T:keywords}. 

We computed Fleiss' $kappa$ on the trial dataset for the five annotators on 21 of the tweets. The value was .83 for Layer A (OFF vs. NOT) indicating high agreement. As to normalization and anonymization, we did not store any user metadata or Twitter IDs, and we substituted the URLs and the Twitter mentions by placeholders.

During the full annotation task, we decided to search for more political keywords as they tend to be richer in offensive content. Thus, we sampled our full dataset, so that 50\% of the tweets come from political keywords, and the other 50\% come from non-political keywords. Within these two groups, tweets were evenly sampled for the keywords. In addition to `gun control', and `to:BreitbartNews' used during the trial annotation, four new political keywords were used to collect tweets for the full dataset: `MAGA', `antifa', `conservatives', and `liberals'. The breakdown of keywords and their offensive content in the full dataset is shown in the bottom of Table~\ref{T:keywords}.

We follow prior work in related areas~\cite{burnap2015cyber,davidson2017automated} and we annotate our data using crowdsourcing. We used Figure Eight\footnote{\url{http://www.figure-eight.com}} and we ensured data quality by (\emph{i})~only hiring annotators who were experienced in the platform, and (\emph{ii})~using test questions to discard annotations by individuals who did not reach a certain threshold. Each instance in the dataset was annotated by multiple annotators and inter-annotator agreement was calculated at the end. 

We first acquired two annotations for each instance. In the case of disagreement, we requested a third annotation, and we then took a majority vote. The annotators were asked to label each tweet at all three levels of the annotation scheme, and we considered there to be agreement only when the annotators agreed on the labels for all levels. Approximately 60\% of the time, the two annotators agreed, and thus no additional annotation was needed. A third annotation was requested for the rest of the tweets; there was no instance when more than three annotations were needed.  

\begin{table}[t]
\centering
\scalebox{0.93}{
\begin{tabular}{cccrr|r}
\toprule
\bf A & \bf B &  \bf C & \bf Training & \bf Test & \bf Total \\ 
\midrule

OFF  &  TIN  &  IND &  $2{,}407$ & $100$ & $2{,}507$\\
OFF  &  TIN  &  OTH &  $395$ & $35$ & $430$ \\
OFF  &  TIN  &  GRP &  $1{,}074$ & $78$ & $1{,}152$\\ 
OFF  &  UNT  &  --- &  $524$ & $27$ & $551$ \\
NOT  &  ---  &  --- &  $8{,}840$ & $620$ & $9{,}460$ \\
\midrule
\bf All &    &     &   $13{,}240$ & $860$ & $14{,}100$ \\
\bottomrule 
\end{tabular}
}
\caption{Distribution of label combinations in OLID.}
\label{tab:labels}
\end{table}

\begin{table*}[!tb]
\centering
\scalebox{0.93}{
\begin{tabular}{l|ccc|ccc|ccc|c}
\toprule
&  \multicolumn{3}{c|}{\bf NOT} & \multicolumn{3}{c|}{\bf OFF} & \multicolumn{3}{c|}{\bf Weighted Average} &  \\ 
\midrule
\bf Model & \bf P &  \bf R & \bf F1  & \bf P &  \bf R & \bf F1 & \bf  P &  \bf R & \bf F1 & \bf  F1 Macro  \\ \hline
SVM & 0.80  & 0.92 &  0.86 & 0.66 &   0.43 &   0.52 & 0.76 &   0.78 &   0.76 & 0.69 \\
BiLSTM    &   0.83  &   0.95  &    0.89  &   0.81  &   0.48   &   0.60 &   0.82  &  0.82   &  0.81 & 0.75   \\
CNN  &  0.87  &    0.93  &  0.90   &    0.78  &  0.63  &  0.70   &  0.82  & 0.82   &  0.81 & \bf 0.80  \\ 
\midrule
All NOT &  - &      0.00    &  0.00 &    0.72    &  1.00    &  0.84 &   0.52    &  0.72     &  0. & 0.42 \\ 
All OFF &  0.28  &   1.00 &   0.44 &  - &   0.00 &  0.00 & 0.08  &    0.28  &    0.12 & 0.22  \\ 
\bottomrule
\end{tabular}
}
\caption{Results for offensive language detection (Level A). We report Precision (P), Recall (R), and F1 for each model/baseline on all classes (NOT, OFF), and weighted averages. Macro-F1 is also listed (best in bold).}
\label{tab:layerA}
\end{table*}

\begin{table*}[!tb]
\centering
\scalebox{0.93}{
\begin{tabular}{l|ccc|ccc|ccc|c}
\toprule
&  \multicolumn{3}{c|}{\bf TIN} & \multicolumn{3}{c|}{\bf UNT} & \multicolumn{3}{c|}{\bf Weighted Average} &  \\ \hline
\bf Model & \bf P &  \bf R & \bf F1  & \bf P &  \bf R & \bf F1 & \bf  P &  \bf R & \bf F1 & \bf  F1 Macro \\ \midrule
SVM & 0.91 &  0.99 &   0.95 &  0.67 &   0.22 &   0.33 &  0.88 &   0.90 &  0.88 & 0.64 \\
BiLSTM   &    0.95    &  0.83   &   0.88   &   0.32    &  0.63  &   0.42 &     0.88    &  0.81  &    0.83  & 0.66 \\
CNN  &   0.94   &   0.90    &  0.92 &   0.32    &  0.63  &   0.42 & 0.88  &   0.86    &  0.87 & \bf 0.69  \\ \midrule
All TIN &  0.89   &   1.00   &   0.94 & - &  0.00 &  0.00 & 0.79  &    0.89 &  0.83 & 0.47 \\ 
All UNT &  -    &  0.00     & 0.00 & 0.11  &    1.00  &    0.20 & 0.01  &    0.11   &   0.02 & 0.10 \\ \bottomrule
\end{tabular}
}
\caption{Results for offensive language categorization (level B).
We report Precision (P), Recall (R), and F1 for each model/baseline on all classes (TIN, UNT), and weighted averages. Macro-F1 is also listed (best in bold).}
\label{tab:layerB}
\end{table*}

\begin{table*}[!tb]
\centering
\scalebox{0.93}{
\begin{tabular}{l|ccc|ccc|ccc|ccc|c}
\toprule
&  \multicolumn{3}{c|}{\bf GRP} & \multicolumn{3}{c|}{\bf IND} & \multicolumn{3}{c|}{\bf OTH} & \multicolumn{3}{c|}{\bf Weighted Average} & \\ \midrule
\bf Model & \bf P &  \bf R & \bf F1  & \bf P &  \bf R & \bf F1 & \bf  P &  \bf R & \bf F1 & \bf  P &  \bf R & \bf F1 &  \bf  F1 Macro  \\ 
\midrule
SVM & 0.66 &  0.50 &  0.57 & 0.61 &   0.92 &   0.73 &  0.33 &   0.03 &   0.05 &  0.58 &   0.62 &  0.56 & 0.45 \\
BiLSTM  &     0.62   &  0.69   &   0.65 &     0.68   &   0.86   &   0.76 &     0.00   &   0.00   &   0.00  &  0.55  &  0.66    & 0.60 & \bf 0.47 \\
CNN  &    0.75   &   0.60  &    0.67 &    0.63   &  0.94  &   0.75 &    0.00   &   0.00  &   0.00  &  0.57  &  0.66    &  0.60 & \bf 0.47  \\ 
\midrule
All GRP & 0.37  &    1.00  &    0.54 &  - &  0.00  &    0.00 & - & 0.00 &    0.00 & 0.13   &   0.37 &     0.20  & 0.18 \\ 
All IND & - &    0.00  &    0.00 & 0.47  &    1.00  &    0.64 & - & 0.00 &      0.00 & 0.22   &   0.47  & 0.30 & 0.21 \\
All OTH & - & 0.00 &      0.00 & - & 0.00 &      0.00 & 0.16  &    1.00   &   0.28 & 0.03  & 0.16 &     0.05 & 0.09 \\ 
\bottomrule
\end{tabular}
}
\caption{Results for offense target identification (level C).
We report Precision (P), Recall (R), and F1 for each model/baseline on all classes (GRP, IND, OTH), and weighted averages. Macro-F1 is also listed (best in bold).}
\label{tab:layerC}
\end{table*}

\noindent The breakdown of the data into training and testing for the labels from each level is shown in Table~\ref{tab:labels}. It is worth noting that one of the key challenges we observed when collecting for OLID was producing a dataset containing a sufficient number of instances for each class. This is particularly evident in the sizes for Subtasks B and C. Other studies also had this issue when collecting similar datasets. For example, in \cite{davidson2017automated}, only 5\% of the tweets were considered hate speech by the majority of the annotators, and in \cite{burnap2015cyber} only 11.6\% of the examples were labeled as hate speech.

\section{Experiments and Evaluation}

We experiment with various models:

\paragraph{SVM} Our simplest machine learning model is a linear SVM trained on word unigrams. SVMs have achieved state-of-the-art results for many text classification tasks  \cite{zampieri2018language}.  

\paragraph{BiLSTM} We also experiment with a bidirectional Long Short-Term-Memory (BiLSTM) model, which we adapted from a pre-existing model for sentiment analysis  \cite{rasooli2018cross}. The model consists of (\emph{i})~an input embedding layer, (\emph{ii})~a bidirectional LSTM layer, and (\emph{iii})~an average pooling layer of input features. The concatenation of the LSTM layer and the average pooling layer is further passed through a dense layer, whose output is ultimately passed through a \textit{softmax} to produce the final prediction. We set two input channels for the input embedding layers: pre-trained FastText embeddings \cite{bojanowski2016enriching}, as well as updatable embeddings learned by the model during training.  

\paragraph{CNN} Finally, we experiment with a Convolutional Neural Network (CNN) model based on the architecture of \cite{kim2014convolutional}, and using the same multi-channel inputs as the above BiLSTM.

\noindent Our models are trained on the training dataset, and evaluated by predicting the labels for the held-out test set. As the label distribution is highly imbalanced (see Table~\ref{tab:labels}), we evaluate and we compare the performance of the different models using macro-averaged F1-score. We further report per-class Precision (P), Recall (R), and F1-score (F1), and weighted average. Finally, we compare the performance of the models against simple majority and minority class baselines.

\subsection{Offensive Language Detection}

The performance on discriminating between offensive (OFF) and non-offensive (NOT) posts is reported in Table~\ref{tab:layerA}.
We can see that all models perform significantly better than chance, with the neural models performing substantially better than the SVM. The CNN outperforms the RNN model, achieving a macro-F1 score of 0.80.

\subsection{Categorization of Offensive Language}

In this set of experiments, the models were trained to discriminate between targeted insults and threats (TIN) and untargeted (UNT) offenses, which generally refer to profanity. The results are shown in Table~\ref{tab:layerB}.  
We can see that the CNN performs better than the BiLSTM, with a macro-F1 score of 0.69. 
Note that all models perform better at identifying TIN compared to UNT.

\subsection{Offensive Language Target Identification}

The results for the offensive target identification experiment are shown in Table~\ref{tab:layerC}. Here the models were trained to distinguish between three targets: a group (GRP), an individual (IND), or others (OTH).
We can see that all three models achieved similar results, far surpassing the random baselines, with a slight performance edge for the neural models.

The performance of all models for the OTH class is 0, which can be explained by two factors. First, unlike the two other classes, OTH is a heterogeneous collection of targets. It includes offensive tweets targeted at organizations, situations, events, etc., thus making it more challenging for models to learn discriminative properties for this class. Second, there are fewer training instances for this class compared to the other two: there are only 395 instances for OTH vs. 1,075 for GRP and 2,407 for IND.

\section{Conclusion and Future Work}

We presented OLID, a new dataset with annotation of type and target of offensive language. It is the official dataset of the shared task \textit{SemEval 2019 Task 6: Identifying and Categorizing Offensive Language in Social Media} (OffensEval) \cite{offenseval}.\footnote{\url{http://competitions.codalab.org/competitions/20011}} In OffensEval, each annotation level in OLID is an independent sub-task. 
To the best of our knowledge, this is the first dataset to contain annotation of type and target of offenses in social media, and it opens interesting research directions. 

We further presented baseline experiments using SVMs and neural networks, which have shown that this is a challenging, yet doable task. 

\noindent In future work, we would like to make a cross-corpus comparison of OLID vs. datasets annotated for similar tasks such as aggression identification \cite{kumar2018benchmarking} and hate speech detection \cite{davidson2017automated}. We further plan to create similar datasets for other languages, following OLID's hierarchical annotation scheme.

\section*{Acknowledgments}

We would like to thank the anonymous NAACL reviewers for their valuable suggestions and Nikola Ljube\v{s}i\'{c} for the feedback provided.

This research presented was partially supported by an ERAS fellowship awarded to Marcos Zampieri by the University of Wolverhampton.

\bibliography{naaclhlt2019}
\bibliographystyle{acl_natbib}

\end{document}